\documentclass{article} %
\usepackage{iclr2027_conference,times}

\usepackage{algorithm}          %
\usepackage{algorithmic}        %
\usepackage{standalone}         %

\usepackage{amsmath,amsfonts,bm}

\def\eqref#1{equation~\ref{#1}}

\def\1{\bm{1}}

\DeclareMathAlphabet{\mathsfit}{\encodingdefault}{\sfdefault}{m}{sl}
\SetMathAlphabet{\mathsfit}{bold}{\encodingdefault}{\sfdefault}{bx}{n}

\usepackage[utf8]{inputenc} %
\usepackage{lipsum} %

\usepackage[T1]{fontenc}

\usepackage{etoolbox}
\newbool{includeappendix}
\setbool{includeappendix}{true}

\ifdefined\isoverfull
\else
\fi

\usepackage{xcolor} %

\definecolor{my-full-blue}{HTML}{1F77B4}

\definecolor{my-full-orange}{HTML}{FF7F0E}

\definecolor{my-full-green}{HTML}{2CA02C}

\definecolor{my-full-red}{HTML}{d62728}

\definecolor{my-full-purple}{HTML}{9467bd}

\colorlet{my-blue}{my-full-blue!30}
\colorlet{my-orange}{my-full-orange!30}
\colorlet{my-green}{my-full-green!30}
\colorlet{my-red}{my-full-red!30}
\colorlet{my-purple}{my-full-purple!30}

\usepackage{listings}

\usepackage{textcomp}

\usepackage{xcolor}

\usepackage[scaled=0.8]{beramono}

\definecolor{ckeyword}{HTML}{7F0055}
\definecolor{ccomment}{HTML}{3F7F5F}
\definecolor{cstring}{HTML}{2A0099}

\lstdefinestyle{numbers}{
	numbers=left,
	framexleftmargin=20pt,
	numberstyle=\tiny,
	firstnumber=auto,
	numbersep=1em,
	xleftmargin=2em
}

\lstdefinestyle{layout}{
	frame=none,
	captionpos=b,
}

\lstdefinestyle{comment-style}{
	morecomment=[l]//,
	morecomment=[s]{/*}{*/},
	commentstyle={\color{ccomment}\itshape},
}

\lstdefinestyle{string-style}{
	morestring=[b]",%
	morestring=[b]',%
	stringstyle={\color{cstring}},
	showstringspaces=false,%
}

\lstdefinestyle{keyword-style}{
	keywordstyle={\ttfamily\bfseries},
	morekeywords={
		function,
		constructor,
		int,
		bool,
		return,
		returns,
		uint
	},
	morekeywords = [2]{},
	keywordstyle = [2]{\text},
	sensitive=true,
}

\lstdefinestyle{input-encoding}{
	inputencoding=utf8,
	extendedchars=true,
	literate=
	{ℝ}{$\reals$}1%
	{→}{$\rightarrow$}1%
	{α}{$\alpha$}1%
	{β}{$\beta$}1%
	{λ}{$\lambda$}1%
	{θ}{$\theta$}1%
	{ϕ}{$\phi$}1%
}

\lstdefinestyle{escaping}{
	moredelim={**[is][\color{blue}]{\%}{\%}},
	escapechar=|,
	mathescape=true
}

\lstdefinestyle{default-style}{
	basicstyle=\fontencoding{T1}\ttfamily\footnotesize,
	style=numbers,
	style=layout,
	style=comment-style,
	style=string-style,
	style=keyword-style,
	style=input-encoding,
	style=escaping,
	tabsize=2,
	upquote=true
}

\lstdefinelanguage{BASIC}{
	language=C++,
	style=default-style
}[keywords,comments,strings]%

\usepackage{booktabs}       %
\usepackage{nicefrac}       %
\usepackage[flushleft]{threeparttable}

\usepackage{url}
\usepackage{xcolor}
\usepackage{xspace}
\usepackage{etoolbox}
\usepackage{fontawesome}
\usepackage{pifont}
\usepackage{enumitem}
\usepackage{amsmath,amssymb}
\usepackage{array}
\usepackage{multirow}
\usepackage{wrapfig}
\usepackage{adjustbox,booktabs}
\usepackage{siunitx}
\usepackage{caption}
\usepackage{fontawesome}
\usepackage{stackengine}
\usepackage{svg}
\usepackage{mathtools}
\usepackage{xspace}
\usepackage{wrapfig}
\usepackage{makecell}
\usepackage{graphicx}
\usepackage{booktabs}
\usepackage{longtable}
\usepackage{hyperref}
\usepackage[labelformat=simple]{subcaption}
\usepackage[most]{tcolorbox}
\usepackage{diagbox}

\usepackage{colortbl}

\usepackage{flafter}

\usepackage{tikz}

\usetikzlibrary{calc,decorations,decorations.pathmorphing}
\usetikzlibrary{positioning,fit,arrows}
\usetikzlibrary{decorations.markings}
\usetikzlibrary{shapes,shapes.geometric}
\usetikzlibrary{shadows,shadows.blur,patterns,snakes}
\usetikzlibrary{backgrounds,decorations.pathreplacing,calligraphy,automata}
\usetikzlibrary{intersections}
\usetikzlibrary{angles,quotes}
\usetikzlibrary{plotmarks}
\usetikzlibrary{patterns}
\usetikzlibrary{arrows.meta}
\usetikzlibrary{shapes.misc}
\usetikzlibrary{chains}
\usetikzlibrary{shapes.callouts}
\usetikzlibrary{shapes.misc}

\usepackage{amsthm}
\usepackage{stmaryrd}

\newcolumntype{x}[2]{S[table-format=#1.#2,table-auto-round]}
\newcolumntype{M}{>{$}l<{$}}

\NewDocumentCommand{\phimodel}{O{}}{%
\textsc{Phi}\ifstrempty{#1}{}{-{#1}}\xspace
}

\newcommand{\gptosssmall}{\textsc{GPT-OSS-20B}\xspace}
\newcommand{\goedelprover}{\textsc{Goedel-Prover-V2}\xspace}
\newcommand{\goedelproversmall}{\textsc{Goedel-Prover-V2-8B}\xspace}
\newcommand{\goedelproverlarge}{\textsc{Goedel-Prover-V2-32B}\xspace}
\newcommand{\goedelformalizer}{\textsc{Goedel-Formalizer-V2-8B}\xspace}

\newcommand{\geminithree}{\textsc{Gemini-3-Pro}\xspace}

\definecolor{acceptblue}{HTML}{6494EA}
\hypersetup{colorlinks=true,linkcolor=acceptblue,citecolor=acceptblue,urlcolor=acceptblue}

\definecolor{lightred}{HTML}{ffcbc7}
\definecolor{gemini}{HTML}{4285F4}
\definecolor{claude}{HTML}{f3e9d7}
\definecolor{deepseek}{HTML}{FADA4B}
\definecolor{qwen}{HTML}{FA574B}
\definecolor{oai}{HTML}{10a37f}
\definecolor{LightGreen}{HTML}{CCFFCC}

\lstdefinestyle{mystyle}{
    breaklines=true,
  language={},                %
  basicstyle=\scriptsize\ttfamily,  
  numbers=none,     %
  columns=fullflexible,       %
  keepspaces=true,            %
  showstringspaces=false,     %
  upquote=true,               %
  mathescape=false,           %
  texcl=false,                %
  escapeinside={},  
  framextopmargin=0pt,
framexbottommargin=0pt,
breakindent=0pt, 
}

\newtcblisting{prompt}[2][]{
    arc=3pt, outer arc=3pt,
    width=\linewidth,
    left=1mm,
    top=0mm,
    bottom=0mm,
    title=#2, 
    colback=oai!5!white,
    colframe=oai!50!black,
    fonttitle=\bfseries,
    listing only, 
    listing options={style=mystyle},
    breakable,
    #1
}

\newtcblisting{system}[2][]{
    arc=3pt, outer arc=3pt,
    width=\linewidth,
    left=1mm,
    top=0mm,
    bottom=0mm,
    title=#2, 
    colback=acceptblue!5!white,
    colframe=acceptblue!50!black,
    fonttitle=\bfseries,
    listing only, 
    listing options={style=mystyle},
    breakable,
    #1
}

\tcbset{
  mybox/.style={
    breakable,
    enhanced,
    colback=white,
    colframe=black!50,
    coltitle=black,
    fonttitle=\bfseries,
    fontupper=\scriptsize,
    sharp corners,
    boxrule=0.5pt,
    left=5pt, right=5pt, top=5pt, bottom=5pt,
  }
}

\definecolor{jsonstring}{HTML}{0B6E99}
\definecolor{jsonnumber}{HTML}{A626A4}
\definecolor{jsonbool}{HTML}{B07D00}
\definecolor{jsonnull}{HTML}{6A737D}
\definecolor{jsonpunct}{HTML}{586069}
\definecolor{jsonbrace}{HTML}{24292E}

\colorlet{punct}{red!60!black}
\definecolor{background}{HTML}{EEEEEE}
\definecolor{delim}{RGB}{20,105,176}
\colorlet{numb}{magenta!60!black}

\lstdefinelanguage{json}{
    basicstyle=\scriptsize\ttfamily,
    numbers=none,
    numberstyle=\scriptsize,
    stepnumber=1,
    numbersep=8pt,
    showstringspaces=false,
    breaklines=true,
    literate=
     *{0}{{{\color{numb}0}}}{1}
      {1}{{{\color{numb}1}}}{1}
      {2}{{{\color{numb}2}}}{1}
      {3}{{{\color{numb}3}}}{1}
      {4}{{{\color{numb}4}}}{1}
      {5}{{{\color{numb}5}}}{1}
      {6}{{{\color{numb}6}}}{1}
      {7}{{{\color{numb}7}}}{1}
      {8}{{{\color{numb}8}}}{1}
      {9}{{{\color{numb}9}}}{1}
      {:}{{{\color{punct}{:}}}}{1}
      {,}{{{\color{punct}{,}}}}{1}
      {\{}{{{\color{delim}{\{}}}}{1}
      {\}}{{{\color{delim}{\}}}}}{1}
      {[}{{{\color{delim}{[}}}}{1}
      {]}{{{\color{delim}{]}}}}{1},
}

\lstdefinestyle{jsonstyle}{
  language=json,
  numbers=none,
  framextopmargin=0pt,
  framexbottommargin=0pt,
  breakindent=2pt
}

\newtcblisting{jsonlisting}[2][]{
  arc=2pt, outer arc=2pt,
  width=\linewidth,
  left=1mm,
  top=0mm,
  bottom=0mm,
  title=#2,
  colback=deepseek!5!white,     %
  colframe=deepseek!50!black,
  fonttitle=\bfseries,
  listing only,
  listing options={style=jsonstyle},
  breakable,
  #1
}

\usepackage[capitalize,noabbrev]{cleveref}

\crefformat{section}{\S#2#1#3}

\crefrangeformat{section}{\S#3#1#4\crefrangeconjunction\S#5#2#6}

\crefmultiformat{section}{\S#2#1#3}{\crefpairconjunction\S#2#1#3}{\crefmiddleconjunction\S#2#1#3}{\creflastconjunction\S#2#1#3}

\newcommand{\crefrangeconjunction}{--}

\crefname{listing}{Lst.}{listings}
\crefname{line}{Lin.}{Lin.}
\crefname{appendix}{App.}{App.}

\newcommand{\app}[1]{%
	\ifbool{includeappendix}{\cref{#1}}{the appendix}%
}
\newcommand{\App}[1]{%
	\ifbool{includeappendix}{\cref{#1}}{The appendix}%
}

\usepackage{hyperref}
\usepackage{url}

\title{Optimizing the Cost-Quality Tradeoff\\
of Agentic Theorem Provers in Lean}

\author{Kári Rögnvaldsson\textsuperscript{*},\space
        Chenhao Sun\textsuperscript{*},\space
        Jasper Dekoninck \& Martin Vechev \\
        ETH Zurich, Switzerland \\
        {\normalfont\textsuperscript{*}Equal Contribution} \\
        \texttt{\small\{kari.roegnvaldsson,chenhao.sun,jasper.dekoninck,martin.vechev\}@inf.ethz.ch}}

\iclrfinalcopy
\begin{document}
\maketitle
\lhead{Preprint}
\suppressfloats[t]

\ificlrfinal
    \vspace{-22pt}
    {\centering
        \raisebox{-0.16em}{\includegraphics[height=1em]{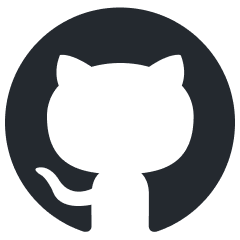}} \href{https://github.com/eth-sri/optimizing-lean-agents}{\nolinkurl{github.com/eth-sri/optimizing-lean-agents}}\par}
    \vspace{4pt}
\fi

\begin{abstract}
Large language models (LLMs) are increasingly used in workflows for generating formal proofs in Lean. These workflows often decompose problems into smaller lemmas, sample many proof attempts, and use compiler feedback to guide search. However, they can be prohibitively expensive, often spending substantial compute on attempts that ultimately fail. In this work, we address this problem with an \emph{action routing agent} that consists of a data plane and a control plane. The data plane generates natural-language lemma decompositions, formalizes them in Lean, and samples proof attempts for the resulting theorem and lemma targets. The control plane observes previous failed Lean attempts, estimates both the likelihood of success and cost of another attempt, and decides whether to continue proving the current target or restart from a new breakdown. On a subset of PutnamBench, our agent decreases the cost by $28.2\%$ over a fixed-step baseline on average, preserving performance while using substantially less compute. These results suggest that failed Lean trajectories provide actionable signals for cost-aware resource allocation in agentic theorem proving.
\end{abstract}

\section{Introduction}

\begin{figure*}[t]
    \centering
    \resizebox{\linewidth}{!}{%
        \providecommand{\lexeme}[1]{\texttt{#1}}%
\definecolor{accentblue}{RGB}{88,139,202}%
\definecolor{lightgreen}{RGB}{220,245,230}%
\definecolor{lightred}{RGB}{250,225,225}%
\definecolor{darkgreen}{RGB}{40,167,69}%
\definecolor{darkred}{RGB}{220,53,69}%
\definecolor{textgray}{RGB}{120,120,120}%
\definecolor{lightgray}{RGB}{240,240,240}%
\definecolor{gray2}{HTML}{FCFCFC}%
\definecolor{overviewblue}{HTML}{347bc6}%
\definecolor{purple}{RGB}{155, 89, 182}%
\definecolor{actionamber}{RGB}{217, 128, 24}%
\definecolor{lightamber}{RGB}{255, 242, 220}%
\begin{tikzpicture}[
    node distance=6mm and 8mm,
    base/.style={
        draw=gray,
        fill=lightgray,
        rounded corners=3pt,
        font=\sffamily,
        blur shadow={shadow xshift=0.5pt, shadow yshift=-0.5pt, shadow opacity=20}
    },
    darkbox/.style={
        draw=gray,
        fill=lightgray!20,
        rounded corners=3pt,
        minimum width=6.3cm,
        minimum height=3.5cm,
        anchor=west,
        xshift=0.3cm
    },
    mainbox/.style={
        base,
        minimum height=4.4cm,
        minimum width=10.7cm,
        align=center
    },
    routerbox/.style={
        base,
        minimum height=4.4cm,
        minimum width=6cm,
        align=center,
        fill=accentblue!20,
        draw=accentblue,
    },
    proposalbox/.style={
        base,
        draw=gray,
        fill=lightgray!20,
        minimum height=0.4cm,
        font=\ttfamily\small,
        minimum width=5cm,
        align=center
    },
    partbox/.style={
        base,
        draw=gray,
        fill=lightgray!20,
        minimum height=0.55cm,
        minimum width=5.5cm,
        align=center,
        text=darkgray,
        font=\small
    },
    actionbox/.style={
        partbox,
        draw=actionamber,
        fill=lightamber,
        text=actionamber!70!black
    },
    smallproposalbox/.style={
        base,
        draw=gray,
        fill=lightgray!20,
        minimum height=0.4cm,
        minimum width=2.3cm,
        align=center,
        font=\ttfamily\scriptsize
    },
    lemmabox/.style={
        base,
        draw=gray,
        fill=lightgray!20,
        minimum height=0.4cm,
        minimum width=1.2cm,
        align=center,
        font=\ttfamily\scriptsize
    },
    smallRouter/.style={
        base,
        draw=actionamber,
        fill=lightamber,
        text=actionamber!70!black,
        minimum height=0.4cm,
        minimum width=1cm,
        align=center,
        font=\ttfamily\scriptsize
    },
    optionbox/.style={
        base,
        minimum width=2.8cm,
        minimum height=0.4cm,
        font=\ttfamily\tiny
    },
    syntaxbox/.style={
        base,
        fill=accentblue!20,
        draw=accentblue,
        minimum width=3cm,
        font=\ttfamily\tiny,
        align=left,
        text width=2.6cm
    },
    rulebox/.style={
        base,
        minimum width=2.8cm,
        minimum height=0.4cm,
        font=\ttfamily\tiny,
        text width=2.1cm
    },
    cfgbox/.style={
        base,
        fill=accentblue!20,
        draw=accentblue,
        minimum width=0.6cm,
        font=\tiny,
        align=center,
        anchor=center,
        text width=0.5cm,
    },
    placeholder/.style={
        fill=gray2,
        rounded corners=1pt,
        inner xsep=1pt,
        inner ysep=1.5pt,
        font=\ttfamily\tiny,
        text height=1ex,
        text depth=0.25ex,
        anchor=base west
    },
    placeholder2/.style={
        fill=gray2,
        rounded corners=1pt,
        inner xsep=1pt,
        inner ysep=1.5pt,
        font=\ttfamily\tiny,
        color=white,
        text height=1ex,
        text depth=0.25ex,
        anchor=base west
    },
    resultbox/.style={
        base,
        minimum height=0.5cm,
        minimum width=1.5cm,
        font=\bfseries,
    },
    accept/.style={resultbox, fill=lightgreen, draw=darkgreen, text=darkgreen},
    smallaccept/.style={resultbox, fill=lightgreen, draw=darkgreen, text=darkgreen, minimum width=1cm, minimum height=0.3cm, font=\small\bfseries},
    smallreject/.style={resultbox, fill=lightred, draw=darkred, text=darkred, minimum width=1cm, minimum height=0.3cm, font=\small\bfseries},
    reject/.style={resultbox, 
        fill=lightred, 
        draw=darkred,
        text=darkred},
    arrow/.style={
        ->,
        thick,
        line width=0.4mm,
        color=overviewblue!80,
        >={Triangle[scale=0.6]}
    }
]

\node[mainbox] (graph) at (0,0) {};

\node[proposalbox, anchor=north, yshift=-0.2cm, xshift=-2.3cm] (problem) at (graph.north) {\textcolor{purple}{theorem} exist\_infinite\_primes};

\node[smallproposalbox, anchor=north east, yshift=-0.4cm, xshift=-0.2cm] (breakdown) at (problem.south) {\emph{Formal Breakdown}};

\node[reject, anchor=north, yshift=-0.8cm] (thmProver1) at (breakdown.south) {\ding{55}};

\node[smallRouter, yshift=-0.6cm, xshift=0.5cm] (routermodule1) at (breakdown.west) {Attempt};

\node[reject, anchor=north, yshift=-0.8cm] (thmProver2) at (thmProver1.south) {\ding{55}};

\node[smallRouter, yshift=-1.1cm] (routermodule2) at (routermodule1.south) {Attempt};

\node[smallRouter, yshift=0.2cm, xshift=1.2cm, minimum width=1cm] (routermodule3) at (thmProver2.east) {Restart};

\node[smallproposalbox, anchor=north west, yshift=-0.4cm, xshift=0.2cm] (breakdown2) at (problem.south) {\emph{Formal Breakdown}};

\node[accept, anchor=north, yshift=-0.8cm] (thmProver3) at (breakdown2.south) {\ding{51}};

\node[smallRouter, yshift=-0.6cm, xshift=0.5cm] (routermodule4) at (breakdown2.west) {Attempt};

\node[lemmabox, anchor=west, xshift=0.9cm, yshift=-0.5cm] (lemma1) at (problem.east) {Lemma 1};

\node[smallreject, anchor=north, yshift=-0.8cm] (reject1) at (lemma1.south) {\ding{55}};

\node[smallRouter, yshift=-0.6cm, xshift=0cm] (routermodule5) at (lemma1.west) {Attempt};

\node[smallaccept, anchor=north, yshift=-0.8cm] (accept1) at (reject1.south) {\ding{51}};

\node[smallRouter, yshift=-0.9cm, anchor=north] (routermodule6) at (routermodule5.south) {Attempt};

\node[lemmabox, anchor=west, xshift=0.25cm] (lemma2) at (lemma1.east) {Lemma 2};

\node[smallaccept, anchor=north, yshift=-0.8cm] (accept2) at (lemma2.south) {\ding{51}};

\node[smallRouter, yshift=-0.6cm, xshift=0cm] (routermodule7) at (lemma2.west) {Attempt};

\node[lemmabox, anchor=west, xshift=0.25cm] (lemma3) at (lemma2.east) {Lemma 3};

\node[smallreject, anchor=north, yshift=-0.8cm] (reject3) at (lemma3.south) {\ding{55}};
\node[smallRouter, yshift=-0.6cm, xshift=0cm] (routermodule8) at (lemma3.west) {Attempt};

\node[smallRouter, yshift=-0.9cm, anchor=north] (routermodule9) at (reject3.south) {Restart};

\draw[arrow] (problem.south -| breakdown.north) -- (breakdown.north);

\draw[arrow] (breakdown.south) -- (thmProver1.north);

\draw[arrow] (thmProver1.south) -- (thmProver2.north);

\draw[arrow] (thmProver2.east) -- ++(0.55,0) |- (breakdown2.west);

\draw[arrow] (breakdown2.south) -- (thmProver3.north);

\draw[arrow] (thmProver3.east) -- ++(0.6,0) -- ++(0,2.1) -| (lemma1.north);
\draw[arrow] (thmProver3.east) -- ++(0.6,0) -- ++(0,2.1) -| (lemma2.north);
\draw[arrow] (thmProver3.east) -- ++(0.6,0) -- ++(0,2.1) -| (lemma3.north);

\draw[arrow] (lemma1.south) -- (reject1.north);
\draw[arrow] (reject1.south) -- (accept1.north);
\draw[arrow] (lemma2.south) -- (accept2.north);
\draw[arrow] (lemma3.south) -- (reject3.north);
\draw[arrow, dotted] (reject3.south) -- (routermodule9.north);

\node[routerbox, anchor=west, xshift=0.3cm] (router) at (graph.east) {};

\node[partbox, anchor=north, yshift=-0.2cm] (attempts) at (router.north) {Agent history $(s,a)$};

\node[partbox, anchor=north, yshift=-0.3cm] (features) at (attempts.south) {Features $x_1 \ldots x_n$};

\node[partbox, anchor=north, yshift=-0.3cm, xshift=-1.5cm, minimum width=2.5cm] (cost) at (features.south) {Cost $\hat{c}$};

\node[partbox, anchor=north, yshift=-0.3cm, xshift=1.5cm, minimum width=2.5cm] (quality) at (features.south) {Quality $\hat{q}$};

\node[partbox, anchor=north, yshift=-1.2cm] (tradeoff) at (features.south) {Tradeoff $\hat{q} - \lambda \hat{c}$};

\node[actionbox, anchor=north, yshift=-0.3cm, xshift=-1.5cm, minimum width=2.5cm] (continue) at (tradeoff.south) {Attempt};

\node[actionbox, anchor=north, yshift=-0.3cm, xshift=1.5cm, minimum width=2.5cm] (stop) at (tradeoff.south) {Restart};

\draw[arrow] (attempts.south) -- (features.north);
\draw[arrow] (features.south) -- (cost.north);
\draw[arrow] (features.south) -- (quality.north);
\draw[arrow] (cost.south) -- (tradeoff.north);
\draw[arrow] (quality.south) -- (tradeoff.north);
\draw[arrow] (tradeoff.south) -- (continue.north);
\draw[arrow] (tradeoff.south) -- (stop.north);

\end{tikzpicture}%
%
    }
    \caption{An overview of our approach. As shown on the left, our agent first decomposes the original problem into formal lemmas. It then attempts to prove the theorem, assuming the lemmas are correct. If the proof fails, the agent router (right) observes the trajectory of past attempts, extracts features, and estimates the cost and quality of future attempts. Based on a cost-quality tradeoff, it then decides whether to continue allocating compute to the current target or create a new breakdown. Once a breakdown is accepted, the agent attempts to prove each lemma, with the same router overseeing the proof trajectory and making dynamic routing decisions at each step.}
    \label{fig:overview}
\vspace{-3mm}
\end{figure*}
\vspace{-1mm}
Large language models (LLMs) have made rapid progress in automated theorem proving with Lean \citep{lean1}, with specialized systems obtaining impressive results on challenging benchmarks \citep{seed-prover, hilbert, numina-lean-agent}. Compared to natural language reasoning, formal theorem proving offers a unique opportunity for LLMs, since the correctness of generated proofs can be automatically verified by a compiler, removing the need for human experts. At the same time, formal theorem proving is extremely challenging because proofs must be syntactically valid and every small step must be rigorously justified. To overcome this difficulty, recent works built agentic workflows that iteratively decompose problems into lemmas, generate multiple proof attempts for each, and self-correct when necessary \citep{hilbert,numina-lean-agent}.

\vspace{-1mm}
\paragraph{Agentic theorem provers are expensive.}  The automatic verifiability of Lean proofs allows these workflows to scale to thousands of attempts per problem, which is critical for achieving high performance on difficult benchmarks. However, they are also prohibitively expensive: \citet{numina-lean-agent} spent $\$50$ per problem to achieve a perfect score on Putnam 2025, and the top performance on PutnamBench reportedly used over $\$40$ per problem \citep{putnambench}. 

\vspace{-1mm}
\paragraph{Limitations of fixed-step policies.}
Existing agents typically scale compute through fixed policies, where every input receives the same pre-specified budgets for decomposition, proof generation, and self-correction. This can waste significant compute on infeasible problems or misformalized statements. The reliance of fixed-step policies thus poses a fundamental efficiency tradeoff: while more attempts improve the probability of success, 
they also incur high costs that may not be justified by the marginal improvement.
Unfortunately, this efficiency perspective has largely been ignored in prior work, which has focused on improving performance rather than optimizing workflow costs.

\vspace{-1mm}
\paragraph{This work: Routing for improved cost-quality tradeoffs.} To mitigate this, we adapt and extend existing works in the routing and cascading literature \citep{zhuge2024languageagentsoptimizablegraphs,aggarwal2025automixautomaticallymixinglanguage,ramirez2024optimisingcallslargelanguage} to the setting of Lean theorem proving with agentic workflows. To do so, we propose a novel agent architecture that uses a lightweight router to dynamically evaluate the expected cost, measured as LLM generation compute, and quality of future proof attempts based on the agent's past trajectory. By learning to identify low-value attempts early, our agent can stop early or restart, significantly reducing unnecessary computation while preserving performance.

\vspace{-1mm}

\paragraph{Agent architecture.}
Our architecture conceptually separates the agent into a proof-generation \textit{data plane} (\cref{fig:overview}, left) and a cost-quality \textit{control plane} (\cref{fig:overview}, right). Inspired by \citet{seed-prover}, the data plane generates lemma-style Lean proofs by decomposing the original problem into natural-language lemmas, formalizing those lemmas in Lean, and generating proof attempts for the resulting theorem and lemma targets. Meanwhile, the control plane observes the prior proof trajectory and uses a lightweight model to estimate both the success probability and the cost of the next attempt. Based on a target cost-quality tradeoff, it then dynamically determines whether allocating further compute is worthwhile.

\vspace{-1mm}
\paragraph{Experimental results.} We evaluate our routing approach on an 85-problem subset of PutnamBench. We find that our dynamic approach significantly outperforms the fixed-step baseline across multiple cost-quality metrics. On average, our agent reduces computational cost by $28.2\%$ and improves solve rate by $7.6\%$ in AUC average over budgets compared with the fixed-step baseline. These results demonstrate that fixed-step policies leave significant room for efficiency improvements and that more research is needed to lower the cost of agentic theorem proving.

\vspace{-1mm}
\paragraph{Key contributions.} Our key contributions are:
\vspace{-2mm}
\begin{itemize}
    \item We identify that fixed-step scaling in Lean agents is critically inefficient, and that reducing unnecessary attempts is key to achieving a better cost-quality tradeoff.
    \item We develop a lemma-style Lean prover equipped with a cost-quality adaptive router (\cref{sec:agent-router}).
    \item We empirically demonstrate that our proposed router successfully decreases wasted compute while maintaining solve rate (\cref{sec:router-experiments}).
\end{itemize}
\vspace{-1mm}

\section{Related Work}\label{sec:related-works}
\vspace{-1mm}
\paragraph{Lean benchmarks.}
To evaluate theorem provers in Lean, researchers have developed several benchmarks. MiniF2F \citep{minif2f} and PutnamBench \citep{putnambench} are two widely used evaluation suites for Lean theorem proving, with PutnamBench focusing on problems from the Putnam  mathematics competition. More recent benchmarks have expanded the scope of evaluation to include research-level problems \citep{dekoninck2026matharena,fate}.

\vspace{-1mm}
\paragraph{Lean prover models.}
To improve capabilities, several efforts have trained specialized models for Lean theorem proving \citep{polu2022formalmathematicsstatementcurriculum, lin2025goedelproverfrontiermodelopensource, wang2025kiminaproverpreviewlargeformal, ren2025deepseekproverv2advancingformalmathematical, goedel-prover-v2}. These models, including \goedelprover \citep{goedel-prover-v2}, use reinforcement learning techniques on large datasets of proof trajectories to improve proof generation and instill reasoning capabilities. They perform very well on easier benchmarks like MiniF2F, but still struggle with harder or more advanced benchmarks.

\paragraph{Agentic prover systems.}
To push the performance barrier, recent systems have adopted agentic workflows that make direct use of the exact verifiability of Lean to scale inference compute significantly. These agents recursively decompose problems into smaller lemmas, sample many proof attempts, and use Lean feedback to guide search \citep{delta-prover, seed-prover, hilbert, numina-lean-agent, Hubert2026AlphaProof, chen2025seedprover15masteringundergraduatelevel, mistral2026leanstral, requena2026minimalagentautomatedtheorem}. They obtain significantly improved solve rates on harder benchmarks, but they also shift the bottleneck toward inference-time compute: many require hundreds of expensive LLM calls per problem, making them prohibitively expensive to run at scale.

\vspace{-1mm}
\paragraph{Routing and cascading.}
Routing and cascading are standard approaches for reducing the cost of LLM systems. Routing selects a model or tool for a query according to a learned policy or predictor \citep{chuang2025learningroutellmsconfidence, ding2024hybridllmcostefficientqualityaware, hari2023tryagerealtimeintelligentrouting}. Usually, a query is routed to a specialized model for a specific topic, or to a more expensive model if the query is predicted to be difficult. Instead, cascading runs a sequence of increasingly expensive LLMs and stops once a response passes a quality criterion \citep{chen2023frugalgptuselargelanguage, ramirez2024optimisingcallslargelanguage}. In open-ended natural language tasks, these quality criteria often rely on proxy signals such as confidence, agreement, or response variance \citep{aggarwal2025automixautomaticallymixinglanguage, jitkrittum2024doesconfidencebasedcascadedeferral}.

\vspace{-1mm}
\paragraph{Routing in agentic environments.}
Recent work extends routing and cascading to multi-agent LLM systems \citep{zhuge2024languageagentsoptimizablegraphs, chen2025optimizingmodelselectioncompound, yue2025masrouterlearningroutellms, su2026difficultyawareagenticorchestrationqueryspecific, kumaresan2026acaradaptivecomplexityrouting}. These methods typically optimize query-level model selection, agent structure, or task allocation across collaborating agents. Our setting differs because we apply routing to the unique environment of agentic theorem proving in Lean, where the router observes structured proof trajectories in an agentic process. To our knowledge, no prior work has explored routing in this setting.

\section{Action Routing Lean Agent} \label{sec:agent-router}
\vspace{-1mm}
Our primary goal is to replace fixed policies with dynamically predicted routing decisions, thereby achieving a better cost-quality tradeoff in Lean agents. This section describes the implementation of our proposed architecture. \cref{sec:data-plane} presents the proof-generation data plane, which operates the lemma-style prover and records the trajectories consumed by the control plane. \cref{sec:control-plane} describes the cost-quality control plane, which evaluates these trajectories to decide dynamically whether the agent should retry the current target or abandon it.
\vspace{-1mm}
\subsection{Proof-Generation Data Plane}\label{sec:data-plane}
The design of the data plane is based on prior work \citep{hilbert,seed-prover}. In particular, it is responsible for decomposing the original problem into tractable subgoals, generating proof attempts for those subgoals, and reconstructing the final proof.  In addition, it produces the attempt trajectories used by the control plane to make routing decisions. The data plane consists of four modules, illustrated in \cref{fig:agent_flow}. Detailed prompts of each module are provided in \cref{app:lemma-style-prover-prompts}.

\begin{figure*}[!t]
\centering

\resizebox{\linewidth}{!}{%
\begin{tikzpicture}[
  >=Stealth,
  node distance=0.5cm %
]

  \definecolor{probBlue}{RGB}{122, 166, 218}
  \definecolor{moduleBlue}{RGB}{48, 113, 169}
  \definecolor{failRed}{RGB}{220, 80, 80}

  \tikzset{
    pipeNode/.style={
      rectangle, rounded corners=4pt, fill=moduleBlue, text=white, 
      align=center, minimum height=1.2cm, minimum width=2.4cm, 
      font=\sffamily\bfseries\small
    }
  }

  \node[pipeNode, fill=probBlue] (p1) {Problem\\Statement};
  \node[pipeNode, right=of p1] (p2) {Breakdown\\Module}; %
  \node[pipeNode, right=of p2] (p3) {Formalizer};
  \node[pipeNode, right=of p3] (p4) {Formal Breakdown\\Prover};
  \node[pipeNode, right=of p4] (p5) {Lemma\\Prover};

  \draw[->, thick, draw=gray!80] (p1) -- (p2);
  \draw[->, thick, draw=gray!80] (p2) -- (p3);
  \draw[->, thick, draw=gray!80] (p3) -- (p4);
  \draw[->, thick, draw=gray!80] (p4) -- (p5);

  \draw[->, thick, draw=gray, rounded corners=6pt]
    ($(p5.east) + (0, 0.2)$) -- ++(0.4, 0) |- ($(p5.north) + (0, 0.3)$)
    node[above, font=\sffamily\small, text=gray!90!black] {Pop from lemma stack}
    -| ($(p5.west) + (-0.4, 0.2)$) -- ($(p5.west) + (0, 0.2)$);

\end{tikzpicture}
}
\caption{Overview of the architecture of the data plane, illustrating the core modular components alongside the recursive lemma dependency resolution.}
\label{fig:agent_flow}
\vspace{-3mm}
\end{figure*}

\vspace{-1mm}
\paragraph{Breakdown module.}
We prompt a general-purpose model (\gptosssmall \citep{gptoss}, with high reasoning effort) to first solve the original problem in natural language. The format of the output breakdown is shown in the breakdown prompt in \cref{app:lemma-style-prover-prompts}. The model organizes the proof into a sequence of lemmas, accompanied by concise proof sketches of each lemma and the main theorem. Lemmas may depend on other lemmas in the same breakdown, so each proof sketch specifies which lemma should be used to prove the current target. If the modules described next fail to produce a full proof for a breakdown, the breakdown is discarded, and the agent generates a new one, up to at most $n_b$ repetitions. If all $n_b$ breakdowns are exhausted, the agent terminates.

\paragraph{Formalizer.}
Next, the agent attempts to formalize the breakdown in Lean. For every natural-language lemma, we sample $n_f$ candidate formalizations from \goedelformalizer \citep{goedel-prover-v2}, compile each candidate against Mathlib, and discard any candidate that produces syntax or type errors. If multiple formalizations of the same lemma compile successfully, we pass them to a general-purpose model (\gptosssmall, with low reasoning effort), together with the original natural-language statement and proof sketch. The model then selects the best formalization.

\paragraph{Formal breakdown prover.}
During proof generation, the agent first attempts to prove the top-level theorem using the formalized lemmas as Lean \texttt{axioms}, which can be used directly to prove the statement. The natural-language proof sketch is included in the prompt to guide the prover. We use \goedelproversmall \citep{goedel-prover-v2} to generate proof attempts for this formal breakdown. In Lean, a proof is successful only if it compiles without errors. If a proof attempt fails to compile, the control plane determines whether to sample another attempt, up to a maximum limit, or to terminate the current breakdown. If the control plane terminates the process, the current breakdown is discarded. If the proof compiles successfully, we extract the subset of lemmas invoked in the proof and push them onto a lemma stack for subsequent proof generation.

\paragraph{Lemma prover.}
Once the formal breakdown has been proven, the agent pops one lemma at a time from the stack and attempts to prove it using the same \goedelproversmall model and control plane logic. To avoid spending compute on unused lemmas, the agent only attempts to prove lemmas that are invoked by an accepted proof of the formal breakdown or by other proven lemmas. The input to the prover includes the formal lemma statement, its natural-language proof sketch, and any dependencies  extracted from the sketch, which are provided as axioms.  If the attempt limit is reached or the control plane decides to stop, the breakdown is discarded. If the lemma is proven successfully, any new lemmas invoked in its proof are pushed onto the lemma stack. This process continues recursively until the stack is empty. Once it is, the problem has been successfully solved.

\vspace{-1mm}

\subsection{Cost-Quality Control Plane}\label{sec:control-plane}

The cost-quality control plane decides whether the agent should continue to prove the current target or terminate the trajectory and move to another candidate breakdown. Its role is to replace a fixed attempt budget with a routing policy that accounts for both the expected benefit and the computational cost of additional proof attempts. In this section, we first formulate the trajectory-level decision problem (\cref{sec:action_space}), then define the cost-quality objective used by the router (\cref{sec:reward_function_policy}), and finally describe the cost and quality estimators used to implement this objective (\cref{sec:router-estimating-cq}).

\vspace{-1mm}

\subsubsection{Trajectory and Action Space}\label{sec:action_space}
We define the object currently being proved as a target, which is either the formal breakdown of the original theorem or an individual lemma produced during decomposition. For a given target, a trajectory consists of the sequence of all failed proof attempts observed so far. Each attempt produces a candidate Lean proof, its compilation result, and auxiliary information such as output length, error messages, and proof structure. The control plane uses this trajectory-level information to decide whether further attempts should be made for the current target or whether the agent should move on to another breakdown. At the beginning of a trajectory, to provide non-trivial information, the agent always samples two proof attempts before the first decision is made.

\vspace{-1mm}
We denote the information available to the router at a decision point by an abstract state $s$. This state summarizes the current failed-attempt trajectory and contains whatever quantities are needed by the cost and quality estimators. The concrete information used by two estimators is described in \cref{sec:router-estimating-cq}.

\vspace{-1mm}

We define the action space of the control plane as $A = \{\text{\textsc{Attempt}}, \text{\textsc{Terminate}}\}$. The action \textsc{Attempt} samples one additional proof for the current target. The action \textsc{Terminate} stops the current trajectory and marks the breakdown as failed. 

\subsubsection{Cost-quality Tradeoff and Routing}\label{sec:reward_function_policy}

Intuitively, the router should continue attempting a target when the next proof attempt has a sufficiently high chance of success, and terminate when the trajectory is unlikely to yield a valid proof.

At each decision point, the router receives the state $s$. Let $\hat{q}(s)$ denote the predicted probability that one additional attempt will solve the current target, which we will refer to as quality, and let $\hat{c}(s)$ denote the predicted cost of generating that attempt. Following the cost-quality routing formulation of \citet{dekoninck2025unifiedapproachroutingcascading}, we define the utility of one additional attempt as
\[
    \tau(s) = \hat{q}(s) - \lambda \hat{c}(s),
\]
where $\lambda > 0$ controls the relative penalty assigned to computational cost. A positive value of $\tau(s)$ indicates that the expected gain in success probability outweighs the cost of another attempt under the chosen budget preference.

The resulting routing policy is
\[
    \pi(s) =
    \begin{cases}
        \text{\textsc{Attempt}}, & \tau(s) > 0,\\
        \text{\textsc{Terminate}}, & \text{otherwise.}
    \end{cases}
\]
Thus, the router continues proving the current target only when the predicted value of an additional attempt is positive. The hyperparameter $\lambda$ traces different operating points along the cost-quality frontier. Smaller values of $\lambda$ make the router more willing to incur high costs, while larger values encourage earlier termination and lower total cost.

\subsubsection{Estimating Cost and Quality}\label{sec:router-estimating-cq}
To instantiate the policy above, the control plane requires two estimates for each decision: the expected cost $\hat{c}(s)$ of another attempt and the success probability $\hat{q}(s)$ of that attempt. We estimate these quantities separately.

\paragraph{Cost estimator.} 
We use the sum of input and output tokens as a proxy for the computational cost of each attempt. We measure cost as LLM generation compute and treat the cost of Lean compilation as negligible. Within a trajectory, the agent generates proof attempts using the same model and the same target. We therefore assume that attempt costs are independent and identically distributed. Under this assumption, the average cost of previous attempts is an unbiased estimator of the cost of the next attempt. Thus, the estimated cost $\hat{c}(s)$ is computed as the average cost of all previous attempts in the trajectory.

\paragraph{Quality estimator.} 
Estimating $\hat{q}(s)$ is more challenging because the value of an additional attempt depends on both the inherent difficulty of the target and the history of previous failures. We design features based on previous failures that may signal whether the prover is making progress or is stuck in a local failure mode. The design of these features is discussed in \cref{sec:feature_engineering}. Using these features, along with the number of attempts made so far, we train a logistic regression model to predict the success probability of the next attempt.

\paragraph{Training the quality estimator.}
To train the estimator, we construct random trajectories on a designated training set of problems using a fixed attempt budget, collecting features at every step from the previous history. Each trajectory terminates either when it reaches the first successful attempt or when the attempt budget is exhausted. The quality estimator is a logistic regression model trained on these trajectories, with the attempt success as the signal and the trajectory features and the number of previous attempts as inputs. The training details are listed in \cref{app:experimental-details}.

\subsubsection{Features for Quality Estimation} \label{sec:feature_engineering}

The quality estimator relies on features extracted from the trajectory of previous proof attempts. These features are designed to capture signals of target difficulty and repeated failure modes. In addition to the number of (failed) attempts in the current trajectory, we use the following features:

\paragraph{Proof similarity.}
We measure how similar the generated proofs are across attempts. To compute this feature, we first normalize each proof by extracting the proof body, renaming variables into a standard format, and removing comments and redundant whitespace. We then compute the average pairwise fuzzy string ratio among all normalized proofs in the trajectory. A low similarity score suggests that the prover is exploring different proof states and failing in different ways, which might indicate that the model is confused and has low intrinsic confidence.

\paragraph{Error diversity.}
We compute the ratio of unique Lean compilation error messages to the total number of compilation errors observed across all attempts. A low diversity ratio indicates that the prover repeatedly encounters the same error, suggesting a persistent failure mode. In contrast, a high diversity ratio suggests that the prover is exploring different strategies and failing in different ways.

\subsection{Extension to Other Lean Agents}\label{sec:extend}

Although we instantiate routing in our own open-source prover, the same control plane principle can be applied to other Lean agents \citep{hilbert,seed-prover}.
The key idea is to separate the agent's proof-generation mechanism from the routing policy that decides which action is optimal. This separation makes the framework extensible to agents with richer action spaces, multiple prover models, or additional self-correction. We do not explore these extensions experimentally due to the high cost of evaluating these agents.

\paragraph{Action space.}
The action space need not be restricted to \textsc{Attempt} and \textsc{Terminate}. For example, an agent could include an action that switches to a stronger prover model when the current model appears to be struggling. Another possible action is self-correction, where the agent uses Lean error messages to repair a failed proof attempt. Such an action may be especially useful when the errors suggest syntactic misuse of Lean rather than a genuine logical gap. If the agent supports recursive decomposition, the action space could also include an action that breaks the current target into smaller subgoals instead of abandoning the breakdown entirely.

\paragraph{Optimal policy.}
With a larger action space, the same cost-quality principle can be generalized by assigning each candidate action an estimated quality gain and cost. The policy then selects the action with the highest predicted utility:
\[
    \pi(s) =
    \arg\max_{a \in A} \hat{q}(a,s) - \lambda \hat{c}(a,s).
\]
This formulation preserves the same routing behavior while allowing the router to choose among multiple proof-generation strategies.

\paragraph{Cost and quality estimation.}
The cost and quality estimators can also be generalized by adapting their inputs to the structure of the agent. When multiple prover models are available, cost can be standardized according to the parameter count. Similarly, the quality estimator can incorporate additional trajectory features that reflect the behavior of new modules. These changes affect only the estimator design and do not change the underlying logic of the control plane.

\ificlrfinal
\begingroup
\setlength{\parskip}{0pt}
\fi
\section{Experiments}\label{sec:router-experiments}

We evaluate our agent on a subset of PutnamBench, focusing on its ability to improve the cost-quality tradeoff of Lean proof generation. Specifically, we test whether our agent can reduce unnecessary proof attempts without substantially lowering the solve rate, compare its performance with a fixed-step agent and an oracle agent that has access to ground-truth success probabilities.

\subsection{Experimental Setup} \label{sec:setup}

We now explain our experimental setup, with additional details provided in \cref{app:experimental-details}. Our empirical results are based on simulating proving of pre-generated proof attempts, using 64 attempts per problem, which allows us to evaluate the routing policy over a large number of seeds and to compute robust cost-quality curves.

\paragraph{Dataset.}
We evaluate on the subset of 85 PutnamBench problems reportedly solved by \goedelproverlarge with pass@192.\footnote{\goedelproverlarge reportedly solves 86 problems \citep{goedel-prover-v2}. However, only 85 are listed in the PutnamBench leaderboard list of solved problems. See the full list \href{https://github.com/trishullab/PutnamBench/blob/b391f48b645c5b6101ed89aca96230226d654b3e/docs/results.json\#L265}{here}.} These 85 problems are randomly split into training and test sets, containing 42 and 43 problems, respectively. The router's quality estimator is trained on the training set, routing results reported in this section are evaluated on the test set.

\ificlrfinal\par\endgroup\fi

\paragraph{Fixed-step baseline.} The fixed-step baseline uses the same data plane pipeline, but replaces adaptive routing with a static policy. Specifically, it allocates a fixed maximum of $k$ proof attempts to each active target before terminating the current breakdown and moving to the next one.

\paragraph{Additional baselines.}
We also evaluate \emph{fixed per-breakdown cost baseline}, which assigns each breakdown a fixed SFLOPs budget and continues the breakdown until that budget is exhausted, and \emph{repeated-error baseline}, which augments the fixed-step policy with early stopping when at least $r$ of the last $n$ generations include the same error. For repeated-error stopping, we sweep the repetition threshold $r$ and window length $n$ from 2 to 64 and choose the set of parameters that have the highest improvement on cost AUC. The best set of parameters on the training set are $r=n=5$, corresponding to five consecutive generations with the same error.

\paragraph{Oracle agent baseline.} 
The goal of our agent is to terminate hard targets early, before spending many failed proof attempts on targets that are unlikely to be solved. Consequently, the gains available from routing depend on the distribution of easy and difficult targets in the evaluation set. We contextualize our results by including an oracle agent that has access to the ground-truth success probability of the current target. This success probability replaces the trajectory features used by our agent, while the router still uses the same logistic quality estimator with the number of previous attempts as an input. The oracle is a privileged information baseline that measures how much our adaptive routing strategy could improve with perfect target-difficulty information.

\paragraph{Evaluation metrics.}
We measure computational cost in scaled floating point operations (SFLOPs), defined as the product of the model's effective parameter count and the number of generated output tokens. We trace cost-solve rate curves by varying $\lambda$ from $2\cdot 10^{-8}$ to $4\cdot 10^{-6}$ for our agent and the oracle, and $k$ over $\{1,2,4,8,16,32,48,64\}$ for the fixed-step baseline, with at most eight breakdowns per problem. We summarize these curves using two area-under-the-curve (AUC) metrics: \emph{cost AUC}, which integrates cost over solve rate (lower is better), and \emph{solve rate AUC}, which integrates solve rate over cost (higher is better). To compute these metrics for routing, we connect every curve to the fixed-step $k=1$ point and add the $k=64$ endpoint ($\lambda=0$ for our agent and the oracle). We then interpolate linearly between points and integrate over the range shared by the methods being compared. By default, we report improvements relative to the fixed-step baseline for routing experiments and whole-proof generation for the data-plane comparison (\cref{sec:data-plane-performance}). For each metric, let $A$ and $A_{\mathrm{ref}}$ denote the AUCs of the method and its reference. The relative reduction in cost AUC, $1-A/A_{\mathrm{ref}}$, measures the savings in average cost at matched solve rates. The relative increase in solve rate AUC, $A/A_{\mathrm{ref}}-1$, measures the gain in average solve rate at matched costs.

\subsection{Main Results} \label{sec:main-results}

\begin{wrapfigure}{R}{0.48\textwidth}
    \vspace{-2\baselineskip}
    \centering
    \includegraphics[width=\linewidth]{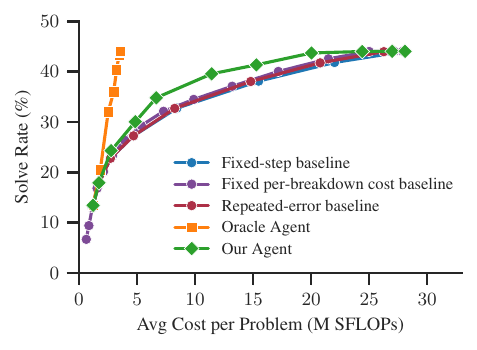}
    \caption{Cost-quality curve of our agent, baselines, and the zero-noise oracle router.}
    \label{fig:real_data_router_cq}
    \vspace{-8mm}
\end{wrapfigure}

\paragraph{Comparison to baselines.}
\cref{fig:real_data_router_cq} compares our agent with the fixed-step baseline and the oracle router. Our agent consistently improves the cost-quality tradeoff relative to the fixed-step baseline, reducing cost AUC by $28.2\%$ and increasing solve rate AUC by $7.6\%$. Compared to the fixed breakdown cost baseline, our agent reduces cost AUC by $22.7\%$ and by $25.8\%$ compared to the repeated-error baseline.

\paragraph{Point comparisons.}
To achieve a high solve rate, the fixed-step baseline solves $44.0\%$ of problems using $28.1$ M SFLOPs, whereas our agent solves $43.7\%$ using $20.0$ M SFLOPs. This corresponds to a $28.7\%$ cost reduction with only a $0.3\%$ decrease in solve rate. In a medium-budget regime, the fixed-step baseline solves $41.8\%$ of problems using $22.0$ M SFLOPs, while our agent solves $41.4\%$ of problems using only $15.3$ M SFLOPs. Relative to these points, our agent reduces cost by $30.6\%$ with only a $0.5\%$ decrease in solve rate.

\paragraph{Contextualizing results.}
The oracle agent achieves a substantially stronger cost-quality tradeoff than both the fixed-step baseline and our agent. This result shows that accurately identifying hard targets is highly valuable for adaptive routing. The gap between our agent and the oracle reference suggests that better trajectory features or stronger quality estimators could further improve the router's ability to terminate low-value targets early. Overall, these results show that fixed-step policies leave significant room for efficiency improvements, and that the routing strategy can eliminate unnecessary proof attempts while preserving most of the solve-rate performance.

\begin{table}[t]
    \centering
    \begin{minipage}[t]{0.48\linewidth}
    \centering
    \caption{Cost AUC reduction relative to the fixed-step baseline across varying levels of artificial noise. Original fixed-split results.}
    \label{tab:noisy_oracle_auc}
    \setlength{\tabcolsep}{3pt}
    \begin{adjustbox}{width=.8\linewidth}
    \begin{tabular}{@{}lc@{}}
        \toprule
        \textbf{Model Configuration} & \makecell{\textbf{AUC Reduction}} \\
        \midrule
        0\% noise (oracle signal) & $71.9\%$ \\
        1\% noise & $57.1\%$ \\
        2\% noise & $38.6\%$ \\
        \textbf{Our agent} & $28.2\%$ \\
        3\% noise & $27.7\%$ \\
        4\% noise & $23.5\%$ \\
        5\% noise & $18.3\%$ \\
        10\% noise & $13.1\%$ \\
        \bottomrule
    \end{tabular}
    \end{adjustbox}
    \end{minipage}\hfill
    \begin{minipage}[t]{0.48\linewidth}
    \centering
    \caption{Cost AUC reduction relative to the fixed-step baseline for different inputs to the quality estimator. Original fixed-split results.}
    \label{tab:feature_ablation_auc}
    \setlength{\tabcolsep}{3pt}
    \begin{adjustbox}{width=.93\linewidth}
    \begin{tabular}{@{}lc@{}}
        \toprule
        \textbf{Model Configuration} & \makecell{\textbf{AUC Reduction}} \\
        \midrule
        \textbf{All three inputs} (our agent) & $28.2\%$ \\
        Proof similarity + error diversity & $22.7\%$ \\
        Error diversity + attempt count & $19.8\%$ \\
        Proof similarity + attempt count & $20.7\%$ \\
        Attempt count only & $13.4\%$ \\
        \bottomrule
    \end{tabular}
    \end{adjustbox}
    \end{minipage}
    \vspace{-4mm}
\end{table}

\subsection{Ablation Studies} \label{sec:ablations}
We now present a series of ablation studies to further analyze the performance of the routing strategy and its components. We first evaluate noisy oracle variants to understand the sensitivity of adaptive routing to quality-estimation accuracy (\cref{sec:noisy-oracle-results}). We then analyze the contribution of different features in the quality estimation model (\cref{sec:feature-ablation}). At the end of this section, we isolate the contribution of the data plane architecture by comparing it against a whole-proof generation approach (\cref{sec:data-plane-performance}).

\subsubsection{Noisy Oracle Agent}\label{sec:noisy-oracle-results}

We evaluate noisy oracle variants by perturbing the ground-truth success probability with different levels of noise. These variants allow us to study how routing performance degrades as the target-difficulty signal becomes less reliable, and to assess how effectively our agent recovers difficulty information from proof trajectories. We perturb the ground-truth success probability of each proof target using different levels of Gaussian or uniform noise. Given a ground-truth success probability of $p$, we refer to a Gaussian noise level of 10\% as $p_n = p + \varepsilon$, where $\varepsilon \sim \mathcal{N}(0, 0.1^2)$, and clamp the value to $[10^{-6},1-10^{-6}]$. We map the perturbed probability $p_n$ with a logit mapping ($\text{logit}(p_n) = \log{p_n/(1-p_n)}$) and feed it as a feature to the quality estimator. \cref{tab:noisy_oracle_auc} reports the cost AUC reduction of each noisy oracle relative to the fixed-step baseline.

The zero-noise oracle, which represents the theoretical ceiling for this estimator family, yields a $71.9\%$ reduction in cost AUC relative to the fixed-step baseline. As the noise level increases, performance gradually decreases. Our agent achieves a $28.2\%$ reduction in cost AUC, placing it between the $3\%$ noise oracle, which achieves $27.7\%$, and the $2\%$ noise oracle, which achieves $38.6\%$. This suggests that the real proof-trajectory features recover a meaningful fraction of the information available to an oracle target-difficulty estimator.

\begin{wrapfigure}[11]{R}{0.45\textwidth}
    \vspace{-2em}
    \centering
    \includegraphics[width=\linewidth]{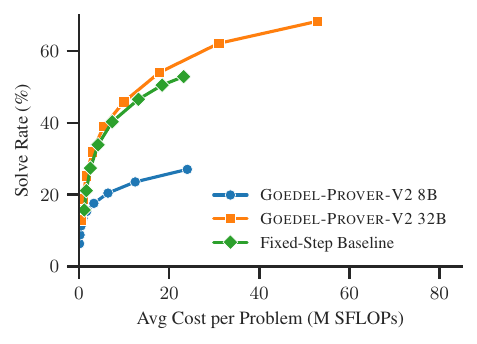}
    \caption{Comparison between the fixed-step data plane and whole-proof generation.}
    \label{fig:comparison_to_baseline_cq}
    \vspace{-8mm}
\end{wrapfigure}

We provide a further analysis of the feature coefficients of the quality estimators in \cref{sec:coefficient-interpretation}.

\subsubsection{Quality Estimator Ablation}\label{sec:feature-ablation}

We next analyze the contribution of different features used by the quality estimator. We compare the full feature set against removing different features. \cref{tab:feature_ablation_auc} reports the cost AUC reduction of each model relative to the fixed-step baseline. As can be clearly seen in \cref{tab:feature_ablation_auc}, all inputs significantly aid our agent in improving the cost-quality tradeoff.

\subsubsection{Data Plane Performance} \label{sec:data-plane-performance}

Finally, we isolate the contribution of the data plane architecture. In addition to the control plane's dynamic routing decisions, the data plane itself substantially improves over a whole-proof generation approach, in which a specialized prover model generates complete proofs without the agentic decomposition architecture. In the whole-proof generation baseline, we prompt the \goedelprover models to generate entire proofs in a single step, report $\text{pass}@k$, with $k$ ranging from 1 to 256. We provide a detailed description of the whole-proof baseline in \cref{app:whole-proof}. This comparison uses all 85 problems and integrates over the data plane's cost range from $k=1$ to $k=64$.

As shown in \cref{fig:comparison_to_baseline_cq}, when using the same prover model \goedelproversmall, the data plane pipeline achieves a substantially higher solve rate: its performance is close to that of the stronger \goedelproverlarge model. Concretely, the data plane pipeline increases solve rate AUC by $91.7\%$ over whole-proof generation with \goedelproversmall, while remaining $6.6\%$ below whole-proof generation with \goedelproverlarge. At the high-budget end, the data plane solves 45 of the 85 problems in our PutnamBench subset with \goedelproversmall as the prover model, up from the 23 solved by the same model under whole-proof generation at $\text{pass}@256$ (\cref{app:whole-proof}). These results demonstrate the effectiveness of the data plane design.

\subsubsection{Routing Model Generalizes Across Agent Configurations} \label{sec:transfer}

\begin{wrapfigure}{R}{0.45\textwidth}
    \centering
    \includegraphics[width=\linewidth]{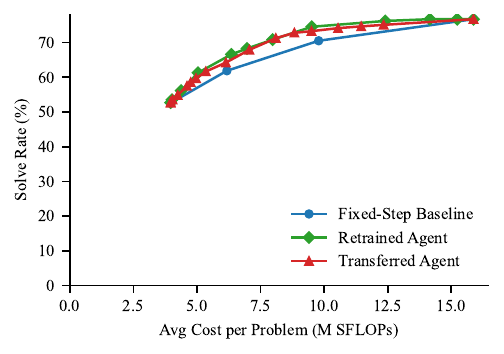}
    \caption{The learned router transfers well to using a different prover model.}
    \label{fig:transfer_cq}
    \vspace{-4mm}
\end{wrapfigure}

We test whether a trained quality estimator generalizes to different models by replacing \goedelproversmall with \goedelproverlarge as the prover and using stronger breakdown and formalizer models. All model replacements are listed in \cref{app:experimental-details}. We compare the \emph{transferred agent}, which reuses the quality estimator trained on the main configuration (\cref{sec:main-results}), with the \emph{retrained agent}, whose quality estimator is trained on data from the new configuration. The transferred estimator reduces cost AUC by $15.2\%$ and increases solve rate AUC by $3.7\%$ relative to the fixed-step baseline (\cref{fig:transfer_cq}). For comparison, the retrained estimator achieves corresponding improvements of $20.0\%$ and $4.8\%$. Our routing model thus generalizes well to the new agent configuration, even without retraining.

\subsubsection{Robustness Across Train-Test Splits and Simulation Seeds}\label{sec:router-robustness}

To assess sensitivity to train-test splits and simulation randomness, we supplement the original fixed-split results with an evaluation over 16 train-test splits of the same 85 problems. We retrain the router on each split's 42 training problems and evaluate on its 43 test problems, calculating the relative AUC improvement of our router compared to the baselines. Each split uses 16 sets of 64 simulation seeds, giving $16 \times 16 = 256$ split-and-seed-set evaluations. \cref{tab:router-robustness-reductions} shows the resulting mean AUC reductions and the confidence intervals of our router compared to each of the baselines. Our agent achieves statistically significant cost AUC reductions against all three baselines (\cref{tab:router-robustness-reductions}).

\begin{table}[t]
    \centering
    \caption{Mean relative cost AUC reductions of our agent and 95\% confidence intervals across splits.}
    \label{tab:router-robustness-reductions}
    \setlength{\tabcolsep}{3pt}
    \begin{tabular}{@{}lcc@{}}
        \toprule
        \textbf{Baseline} & \textbf{Mean router cost AUC reduction} & \textbf{95\% CI} \\
        \midrule
        Fixed-step baseline & $24.1\%$ & $[19.0\%, 29.2\%]$ \\
        Fixed per-breakdown cost baseline & $18.8\%$ & $[14.7\%, 22.9\%]$ \\
        Repeated-error baseline & $22.1\%$ & $[17.4\%, 26.7\%]$ \\
        \bottomrule
    \end{tabular}
\end{table}

\section{Limitations and Future Work}\label{sec:limitations-future-work}
This work studies adaptive routing as a first step toward cost-aware control for agentic formal mathematical proving. In this section, we discuss several limitations of the current system and open directions for future work.

\paragraph{Small benchmark size.}
Our evaluation is limited to a subset of PutnamBench because running the full agent on larger benchmarks is computationally infeasible. We estimate our runs on the 85 problem subset to have cost around 1600-2000 GPU hours on NVIDIA GH200, meaning that if we were to scale to the full PutnamBench, the cost would be roughly eightfold, around 12000-16000 GPU hours. However, this subset is well matched to the capability of the underlying prover and thus provides a useful setting for studying routing behavior.

\paragraph{Gap to oracle routing.}
The proposed routing system also leaves a substantial gap to the oracle router. This gap suggests that the current cost and quality estimators capture only part of the information needed for effective routing. New reward formulations, better trajectory features, and stronger estimator architectures may further improve the prediction.

\paragraph{Limited action space.}
The current implementation is also restricted to a binary action space: the router decides whether to sample another proof attempt for the current target or terminate it. As discussed in \cref{sec:extend}, the same framework could be extended to larger action spaces and more general agentic architectures. In such settings, however, routing becomes more difficult because the agent must decide not only whether to continue, but also which action to take next. Future work could study how to effectively implement adaptive routing for more sophisticated Lean agents with actions such as model switching, self-correction, and recursive decomposition.

\section{Conclusions}\label{sec:conclusions}
In this work, we introduced an action routing agent that uses failed proof trajectories to estimate the cost and marginal value of an additional proof attempt. The agent then uses these estimates to decide whether to continue proving the current target or terminate the current trajectory and restart from a new breakdown. On a subset of PutnamBench, our agent reduces computational cost on average by $28.2\%$ and improves solve rate by $7.6\%$  in AUC average over budgets compared with the fixed-step baseline. Overall, these results suggest that cost-aware control policies are a promising approach for making formal theorem proving more efficient. Instead of optimizing efficiency of the underlying  prover or the number of tokens used by each model call, we show that optimizing on the action level can also lead to significant improvements.

\message{^^JLASTBODYPAGE \thepage^^J}

\subsection*{AI Use Statement}
In this work, we used generative AI tools to aid and polish writing, to support qualitative data analysis, and to interpret results. We have not used generative AI tools for any other required disclosure task, and the rest of the required disclosure tasks are not applicable to this work. Additionally, we used generative AI tools for code generation and to polish writing. 

We have extensively reviewed all AI-assisted work. This includes reviewing and editing all text generated by AI and verifying the correctness of all code generated by two authors through extensive testing and manual verification of code and model outputs. Qualitative conclusions extracted by AI were manually verified and checked for correctness. No AI was used to create any of the main ideas or scientific contributions.  We take responsibility for the final content of this work, including text, claims or artifacts produced with the aid of generative AI.

\subsection*{Ethics Statement}
This research does not involve human subjects or sensitive data. The work focuses on improving the efficiency and effectiveness of theorem proving in Lean, which is a mathematical proof assistant. The methods and results presented in this paper are intended to advance the field of automated theorem proving and do not raise any ethical concerns related to privacy, consent, or potential harm to individuals or communities.

\subsection*{Reproducibility Statement}
We describe our method in detail in this paper, and provide our \href{https://github.com/eth-sri/optimizing-lean-agents}{code on GitHub} for reproducing our results. This repository also includes \href{https://github.com/eth-sri/optimizing-lean-agents/blob/main/proof_simulation/README.md}{detailed README instructions} for running the experiments, along with a dashboard to view and analyze the results. As the creators of PutnamBench explicitly prohibit distribution of formal proofs of the problems, we do not include the proofs generated by our agent in the repository.

\ificlrfinal
    \subsubsection*{Acknowledgements}
This work was supported by a grant from the Swiss National Supercomputing Centre (CSCS) under project ID a155 on Alps.

\fi

\bibliography{references}
\bibliographystyle{iclr2027_conference}

\message{^^JLASTREFERENCESPAGE \thepage^^J}

\ifbool{includeappendix}{%
	\clearpage
	\appendix
	\crefalias{section}{appendix}
	
\section{Experimental Details}\label{app:experimental-details}

\paragraph{Training data.}
To ensure our experiments are non-contaminated, we split the 85 problem subset of PutnamBench (see \cref{sec:setup}) into a training set of 42 problems and a test set of 43 problems. The training set is used to train the quality estimator, while all routing results reported in \cref{sec:router-experiments} are evaluated on the test set. The train-test split is randomly generated and can be found in the \href{https://github.com/eth-sri/optimizing-lean-agents/blob/main/dataset/putnam_rewrite_solved_train_test_split.txt}{GitHub repository}.

\paragraph{Quality estimator.}
The quality estimator is a single logistic regression model trained on trajectories with attempt success as the binary target. Its input consists of two parts: the trajectory features described in \cref{sec:feature_engineering}, and an attempt-count feature $1/m$, where $m$ is the number of previous attempts on the current target. We standardize all inputs to zero mean and unit variance with \texttt{scikit-learn}'s \texttt{StandardScaler} \citep{pedregosa2018scikitlearnmachinelearningpython} and use its logistic regression with defaults.

\paragraph{Monte Carlo simulation.} To obtain a robust estimate of the cost-quality curve for a given $\lambda$, we first pre-generate all proof attempts, sampling 64 attempts per target. We then run Monte Carlo simulations over 64 seeds of the routing policy on each problem, allowing at most eight breakdowns per problem. In each simulation, the environment reshuffles the order of the breakdowns and of the pre-generated attempts, so that the routing policy is evaluated over a diverse set of trajectories. We apply the same procedure to the fixed-step baseline, replacing the cost penalty $\lambda$ with the per-target attempt budget $k$. For the oracle router, the ground-truth success rate of a target is the fraction of successful attempts among its 64 generated attempts.

For each breakdown, we pre-generate 64 proof attempts for the main theorem and each formalized lemma. During replay, the accepted theorem-level proof determines which lemmas enter the proof stack. Attempts are then sampled from the corresponding lemma pools, and successful lemma proofs can add further required lemmas. Thus, the targets visited depend on the successful proofs.

\paragraph{Routing hyperparameters.}
We sweep the cost penalty $\lambda$ (\cref{sec:reward_function_policy}) over 64 seeds with at most eight breakdowns per problem over
\[
    \{2 \cdot 10^{-8},\,4 \cdot 10^{-8},\,6 \cdot 10^{-8},\,8 \cdot 10^{-8},\,1\cdot 10^{-7},\,1.5\cdot 10^{-7},\,2\cdot 10^{-7},\,4\cdot 10^{-7},\, 4\cdot 10^{-6}\},
\]
where $\lambda = 0$ always continues the current target. The fixed-step baseline sweeps the per-target attempt budget $k \in \{1,2,4,8,16,32,48,64\}$ with at most eight breakdowns per problem.

\paragraph{Per-breakdown cost baseline.} We sweep 19 per-breakdown caps in output SFLOPs. The caps were $B_j = \mathrm{round}(2m \cdot 2^{j/2})$ for \(j = 0, \ldots, 18\), where \(m\) is the median output SFLOP cost of failed initial 8B attempts on the training set. For our original train-test split, we have $m = 41968$.

\paragraph{Routing cost model.}
On the original fixed train-test split, using the sum of output and input tokens in the routing cost estimate improves cost-AUC reduction from $25.0\%$ to $28.2\%$ compared with an output-only estimate, with the quality estimator, $\lambda$ grid, seeds, and evaluation protocol fixed. Reported cost uses output-token SFLOPs in both cases, as this is a better estimate of the actual computational cost.

\begin{table}[t]
    \centering
    \caption{Learned logistic regression coefficients for the quality estimator trained on the original fixed train--test split. The model parameters are scaled to mean 0 and variance 1.}
    \begin{tabular}{lr}
        \toprule
        \textbf{Feature} & \textbf{Our Agent} \\
        \midrule
        Intercept & $-4.85$ \\
        Proof Similarity & $0.28$ \\
        Error Diversity & $1.02$ \\
        Attempt Count ($1/x$ mapped) & $0.35$ \\
        \bottomrule
    \end{tabular}
    \label{tab:router-model-params}
\end{table}

\paragraph{Large model experiments.}
In the transfer experiment (\cref{sec:transfer}), we replace \goedelproversmall with \goedelproverlarge as the prover, \gptosssmall with \geminithree as the breakdown model, and \goedelformalizer with \textsc{Goedel-Formalizer-V2-32B} as the formalizer. The per-target attempt limit decreases from 64 to 16. We use the same 85 problems and the original fixed train-test split shown in \cref{fig:real_data_router_cq}, keeping the other settings from \cref{sec:setup}. The transferred agent reuses the quality estimator trained on the main configuration without retraining. The retrained agent fits the same estimator on the new configuration's training problems. Both agents take two initial attempts per target before routing.

For all methods in this comparison, reported SFLOPs exclude the cost of \geminithree breakdown generation because its parameter count is unavailable. The fixed-step baseline sweeps $k \in \{2,4,8,16\}$, with AUCs computed between its $k=2$ and $k=16$ points. To cover this cost range, we sweep $\lambda$ over
\[
    \left\{\begin{aligned}
        &0,\,5\cdot 10^{-9},\,10^{-8},\,2\cdot 10^{-8},\,4\cdot 10^{-8},\,6\cdot 10^{-8},\\
        &8\cdot 10^{-8},\,10^{-7},\,2\cdot 10^{-7},\,10^{-6},\,4\cdot 10^{-6}
    \end{aligned}\right\}
\]
for both routing agents. For the transferred agent, we extend the sweep with $\lambda \in \{2.5\cdot 10^{-10},\,5\cdot 10^{-10},\,10^{-9},\,2\cdot 10^{-9},\,3\cdot 10^{-9}\}$. The transferred estimator, trained with the weaker prover, predicts lower success probabilities, so smaller values of $\lambda$ are needed to reach the high end of the cost range.

\paragraph{Language models.}
The prover \goedelproversmall, formalizer \goedelformalizer, and general-purpose model \gptosssmall all use temperature $1.0$. The two 8B models additionally use $\text{top-}p = 0.95$. We set the maximum generation length to $40960$ tokens for the 8B models and $81920$ for \gptosssmall.

\section{Quality Estimator Analysis}\label{sec:coefficient-interpretation}

The learned coefficients of the quality estimator, shown in \cref{tab:router-model-params}, offer some interesting insights. As expected, the model predicts a lower success probability as more attempts accumulate on a target. Both proof similarity and error diversity correlate positively with success: the model favors attempts whose proofs are similar to each other but whose errors are diverse. This seems paradoxical at first. We hypothesize that a shared high-level structure across proofs (high proof similarity) signals that the model is confident in the overall proof strategy, while the diversity of errors reflects variation in low-level details rather than a single recurring mistake. If every attempt fails in the same way, the model is stuck and unlikely to solve the problem.

\section{Whole-Proof Generation} \label{app:whole-proof}

To analyze and later optimize the cost-quality tradeoff of Lean prover models, it is insightful to first analyze the cost-quality tradeoff of the open-weight \goedelprover models. These results serve as a baseline for our other experiments.

\subsection{Experimental Setup}\label{sec:whole-proof-experiments}
We set up an experiment on the same subset of PutnamBench as we used in the main experiment, attempting each problem 256 times, and examining the success rate of the 8B and 32B models.

To create a robust estimate of the cost-quality curve per model, for each problem, we shuffle the order of the attempts and loop through them sequentially one by one, until we have generated a successful attempt or we have reached the maximum number of attempts. To compute the cost-quality curve of the model, we report $pass@k$ for varying $k$ from 1 to 256, running 64 seeds per $k$ to create stable measurements.

\subsection{Results}

The cost-quality curve of the two models on PutnamBench can be seen in \cref{fig:whole-proof-putnam-cq}. We see that for the PutnamBench, \goedelproverlarge outperforms \goedelproversmall on the entire data set in terms of the cost-quality tradeoff. At 6.4 M SFLOPs per problem, \goedelproversmall solves $20.5\%$ of problems, while at 5.5 M SFLOPs per problem, \goedelproverlarge solves $39.0\%$, showing nearly double performance at a lower cost. We note that \citet{goedel-prover-v2} report 25 problems solved by \goedelproversmall at $\text{pass}@32$ under whole-proof generation, compared to the 23 we observe at $\text{pass}@256$ on our 85-problem subset of PutnamBench.

\begin{figure}[t]
    \centering
    \includegraphics[width=0.8\linewidth]{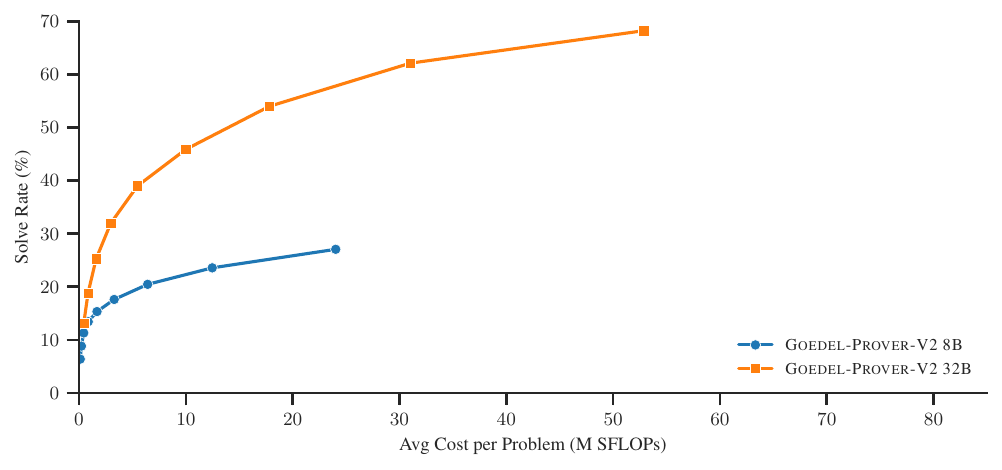}
    \caption{Whole-proof cost-quality curves on all 85 problems in the PutnamBench subset.}
    \label{fig:whole-proof-putnam-cq}
\end{figure}

\section{Data Plane Prompts}\label{app:lemma-style-prover-prompts}
This section lists the prompts used for the different components of the agentic prover system described in \cref{sec:data-plane}.

\tcbset{before skip=6pt, after skip=6pt}

\begin{prompt}{Breakdown Module Prompt}
Break the following Lean problem down into lemmas, then describe how to combine the lemmas into a full solution:

{formal_statement}

**Important points**

1. Make every lemma **mathematically correct** and easy to prove.
2. Do not create redundant lemmas or lemmas that restate the assumptions of the problem
3. In every lemma, also present everything we need to assume to prove the lemma in an isolated manner
4. Make the final solution that combines the lemmas as trivial as possible, i.e. try to put all logic into the lemmas

The concrete breakdown format should be a valid json. The output format and what you need to include is the following:

```json
{{
    "lemmas": [
        {{
            "id": <lemma number>,
            "statement": <lemma statement>,
            "assumption": <state the necessary assumptions for the lemma>,
            "proof": <idea of the proof of the lemma in natural language, if you need to use other lemmas, specify it in the proof idea>
        }}
    ],
    "theorem": {{
        "statement": <repeat the problem statement>,
        "proof": <idea of the proof, how to combine the lemmas into the final solution>
    }}
}}
```

**Important notice:** In lemmas' and theorem's proof ideas, explicitly specify them if you need to use other lemmas' conclusion.

The output can only contain the json above, no other content is allowed.
\end{prompt}

\begin{prompt}{Formalizer Prompt}
This is an informal breakdown of the math problem we have at hand:

{informal_breakdown}

Your task is to formalize Lemma {i}. To reiterate, here is the statement, the assumptions and the proof idea of Lemma {i}:

**Statement:** {statement}

**Assumptions:** {assumption}

**Proof Idea:** {proof}

The statement should have the following format. Please only state lemma {i}, DO NOT try to formalize the final result of the problem.

```lean4
theorem lemma{i} : ... := by sorry
```
\end{prompt}

\begin{prompt}{Formalization Selection Prompt}
Here is a mathematical problem:

**Problem Statement:** {original_statement}

An LLM has been prompted to formalize the following lemma which is supposed to be a substep of the theorem:

**Statement:** {statement}

**Assumptions:** {assumption}

**Proof Idea:** {proof}

Here are four separate formalizations of the lemma:

```
{lemma_code}
```

Choose the best matching formalization to the lemma according to the statement and the assumptions.

Output ONLY the index number (zero-based) of the best formalization, nothing else. Do not include any explanation after the number.

For example, if formalization 2 is best, output only:
```
2
```
\end{prompt}

\begin{prompt}{Theorem/Lemma Prover Prompt}
Complete the following Lean 4 code:

```lean4
{formal_statement}
```

Before producing the Lean 4 code to formally prove the given theorem, give an overview of the lemmas and how they can contribute to solving the problem. The lemmas mentioned correspond to the axioms and should be assumed proven (they will be proven in subsequent steps so don't worry about them being marked as axioms). Restate the proof idea and possibly develop some underdeveloped steps. Here is the proof idea:

{informal_prefix}

**Very Important:** The lemmas should be combined in the way the proof idea suggests. Trust the provided solution.

Your answer must contain the whole Lean 4 proof of the problem in a code block in the following format:

```lean4
import Mathlib

{formal_statement}
<Your proof here>
```
\end{prompt}

}{}

\message{^^JLASTPAGE \thepage^^J}

\end{document}